\documentclass[journal]{IEEEtran}
\usepackage{amssymb}
\usepackage{amsmath,graphicx,bm,threeparttable,indentfirst,cite}
\usepackage{booktabs}
\usepackage{color, multirow,graphicx}
\usepackage{multirow,epsfig,fbox,amsfonts,amsmath,multicol,enumitem,color,pifont}
\usepackage{hyperref}
\usepackage{color}
\usepackage{float,pifont}

\usepackage{tabularx}

\usepackage[linesnumbered,ruled,vlined]{algorithm2e}
\SetKwInput{KwIn}{Input}
\SetKwInput{KwOut}{Output}
\SetKwComment{Comment}{/* }{ */}
\usepackage{tabularx}

\ifCLASSINFOpdf
\else
\fi
\begin{document}
%
\title{FedEU: Evidential Uncertainty-Driven Federated Fine-Tuning of Vision Foundation Models for Remote Sensing Image Segmentation}
%
%
%

 \author{Xiaokang~Zhang,~\IEEEmembership{Senior Member,~IEEE,} Xuran~Xiong,~Jianzhong~Huang~and~Lefei~Zhang,~\IEEEmembership{Senior Member,~IEEE}

\thanks{This work was supported in part by the National Natural Science Foundation of China under Grant No. 42371374.}
\thanks{Xiaokang Zhang is with the School of Artificial Intelligence, Wuhan
University, Wuhan 430072, China (e-mail: zhangxiaokang@whu.edu.cn)}
\thanks{Xuran Xiong is with the School of Information Science and Engineering, Wuhan University of Science and Technology, Wuhan 430081, China (e-mail:  xiongxuran@wust.edu.cn).}
\thanks{Jianzhong Huang and Lefei Zhang are with the School of Computer Science, Wuhan
University, Wuhan 430072, China (email: hjz$\_$2000@whu.edu.cn; zhanglefei@whu.edu.cn).}
\thanks{Manuscript received April 19, 2005; revised August 26, 2015.}}


%
%

\markboth{Journal of \LaTeX\ Class Files,~Vol.~14, No.~8, August~2015}%
{Shell \MakeLowercase{\textit{et al.}}: Bare Demo of IEEEtran.cls for IEEE Journals}
%



\maketitle



\begin{abstract}
Remote sensing image segmentation (RSIS) in federated environments has gained increasing attention because it enables collaborative model training across distributed datasets without sharing raw imagery or annotations. Federated RSIS combined with parameter-efficient fine-tuning (PEFT) can unleash the generalization power of pretrained foundation models for real-world applications, with minimal parameter aggregation and communication overhead.
However, the dynamic adaptation of pretrained models to heterogeneous client data inevitably increases update uncertainty and compromises the reliability of collaborative optimization due to the lack of uncertainty estimation for each local model. To bridge this gap, we present FedEU, a federated optimization framework for fine-tuning RSIS models driven by evidential uncertainty. Specifically, personalized evidential uncertainty modeling is introduced to quantify epistemic variations of local models and identify high-risk areas under local data distributions. Furthermore, the client-specific feature embedding (CFE) is exploited to enhance channel-aware feature representation while preserving client-specific properties through personalized attention and an element-aware parameter update approach. These uncertainty estimates are uploaded to the server to enable adaptive global aggregation via a Top‑$k$ uncertainty-guided weighting (TUW) strategy, which mitigates the impact of distribution shifts and unreliable updates. Extensive experiments on three large-scale heterogeneous datasets demonstrate the superior performance of FedEU. 
More importantly, FedEU enables balanced model adaptation across diverse clients by explicitly reducing prediction uncertainty, resulting in more robust and reliable federated outcomes.
The source codes will be available at \href{https://github.com/zxk688/FedEU}{https://github.com/zxk688/FedEU}.
\end{abstract}
\begin{IEEEkeywords}
Image segmentation, federated learning, fine-tuning,  evidential learning, 
and remote sensing imagery.
\end{IEEEkeywords}

\IEEEpeerreviewmaketitle

\section{Introduction}
\IEEEPARstart{R}{emote} sensing image segmentation (RSIS) plays a critical role in numerous earth observation applications, including land cover/use mapping, disaster assessments, urban management, and environmental monitoring. With the growing availability of high-resolution remote sensing imagery, it has been widely used for pixel-wise identification and segmentation of specific objects, such as buildings, roads, vegetation, and water bodies, from satellite or aerial imagery \cite{9653801,zhang2021stagewise}. In this context, deep learning models such as Convolutional Neural Networks (CNNs) and Transformers have become dominant, delivering high accuracy and efficiency across complex remote sensing datasets \cite{zhang2022artificial}. Recent works leveraging the power of pretrained foundation models with millions (or billions) of parameters, represented by Transformers, have shown promising results when fine-tuned for real-world remote sensing applications \cite{10637992,10485462}. 
Through large-scale pretraining on extensive datasets, these models learn universal representations that enable efficient transfer to downstream tasks \cite{fuller2024croma,bastani2023satlaspretrain}.

Traditional deep learning methods typically rely on centralized training, where the local data from participants is sent to a central data processing center for processing. Unfortunately, remote sensing datasets are often dispersed across institutions due to concerns over geospatial security, storage limits, and industrial competition, resulting in the widespread issue of data islands  \cite{li2025unleashing}. In real applications and engineering, distributed computing has been successfully applied, especially when large volumes of satellite imagery are available \cite{tan2025towards,10812017,9244062}. While this approach enables direct access to large datasets and straightforward implementation, it poses serious privacy concerns when sensitive or proprietary data, such as ultra-high-resolution images and annotations from different clients, are inevitably involved \cite{moreno2024federated,xu2023ai}.

Recently, federated learning (FL) has gained attention as an effective approach to tackling this issue by facilitating collaborative model development across multiple parties while keeping data decentralized. This paradigm preserves data privacy and mitigates security risks by avoiding the transmission of raw data. For RSIS tasks, FL enables institutions to jointly train powerful models without sharing imagery or annotations, thereby breaking data barriers, improving generalization across regions, and advancing scalable remote sensing interpretation \cite{tan2024bridging,10596122}. Unfortunately, applying traditional FL directly to large pretrained models is impractical, as the optimization and transmission of billions of parameters incur substantial computational burdens on clients and lead to high communication overhead. To address this limitation, parameter-efficient fine-tuning (PEFT) methods provide an effective alternative by restricting optimization and communication to only a small subset of parameters \cite{10857375,10637992, 10849617}. 
Moreover, embedding efficient fine-tuning of pretrained models within a federated learning framework can enhance the efficiency of adopting foundation models for vision tasks in distributed and privacy-sensitive demands, while reducing communication overhead between the server and clients.

Existing federated optimization studies have predominantly explored FedAvg \cite{pmlr-v54-mcmahan17a} as the baseline approach, where each client trains locally on its own dataset and sends model updates to a central server for averaging aggregation. To tackle the model drifts caused by data heterogeneity, various local regularization and enhanced global aggregation strategies have been investigated \cite{hao2025fedcs,wang2024fedsls}. Building on this, personalized federated optimization methods have been proposed to accommodate client-specific variations while maintaining shared knowledge, by dynamically adapting updates based on local data or jointly optimizing global and local models \cite{tan2022towards,sabah2024model,10445525,NEURIPS2020_24389bfe,zhang2023fedala,10345648}. Despite those achievements, federated fine-tuning for RSIS tasks still encounters several key challenges. 
\begin{enumerate}
  \item Most FL methods focus on image-level recognition and classification tasks \cite{10988823,hu2025heterogeneity,10496080,10858749,9926002}. 
  For RSIS tasks that require pixel-wise prediction, spatial and statistical heterogeneity often lead to inconsistent optimization objectives among clients \cite{tan2024bridging,10596122}, highlighting the need for specialized federated learning strategies. 
  \item In federated optimization, balancing global generalization and local personalization remains difficult \cite{zhang2024improving,10605121}. Unreliable local predictions and biased updates can mislead global aggregation, degrading performance and causing unstable convergence. Without reliable uncertainty estimation, abnormal updates cannot be identified or suppressed, further contaminating the global model.
    \item Despite its potential, federated fine-tuning of a pretrained model is still underexplored for RSIS. Existing methods mainly employ a simple combination of centralized PEFT and the FedAvg algorithm \cite{chen2024feddat,10095356,10210127,Liu_FedFMS_MICCAI2024,10.1007/978-3-031-77610-6_21}. 
    Unlike conventional FL, federated fine-tuning must dynamically adapt pretrained models to heterogeneous client data while maintaining global consistency \cite{m2024personalized}. This process amplifies update uncertainty and compromises the reliability of collaborative optimization, making it unclear whether federated fine-tuning can deliver stable and trustworthy gains over local adaptation.
\end{enumerate}

To bridge these gaps, we proposed an evidential uncertainty-driven federated fine-tuning framework named FedEU for RSIS based on evidential deep learning. 
A key component is the computation of the epistemic uncertainties associated with each client’s predictions via a domain-specific evidential uncertainty head, which captures the evidence-based confidence of the shared model for the client’s data distribution.  By uploading the uncertainties and updated model parameters to the server, the framework dynamically adjusts the aggregation process to optimize global model updates based on local models' prediction uncertainty.
The main contributions of this article are as follows.
\begin{itemize}
\item {We propose a novel evidential uncertainty-driven personalized federated fine-tuning framework, where epistemic uncertainty is fully exploited to guide both local training and global aggregation. To the best of our knowledge, this is the first work introducing evidential learning into federated optimization.}
\item We design two personalization strategies: personalized evidential estimation for uncertainty-guided local training, and client-specific feature embeddings (CFE) with adaptive element-wise parameter update, improving robustness to spatial heterogeneity and imaging variations.
\item A Top-$k$ uncertainty-guided weighting (TUW) strategy is developed for global aggregation, which identifies high-risk regions, down-weights unreliable models, and suppresses misleading local patterns to dynamically optimize the aggregation process.
\item Extensive experiments on three large-scale datasets demonstrate that our approach achieves significant improvements in prediction performance and robustness for federated RSIS.
\end{itemize}

The remainder of this article is structured in the following manner.  Section \ref{sec2} provides a concise review of related works on federated fine-tuning for the RSIS task. Section \ref{sec3} outlines the key components and training process of the proposed FedEU approach. Section \ref{sec4} presents the performance evaluation and comparative analysis of FedEU. Finally, Section \ref{sec6} concludes the study.

\section{Related works}\label{sec2}

\begin{figure*}[htp]
	\centering
\includegraphics[width=\linewidth]{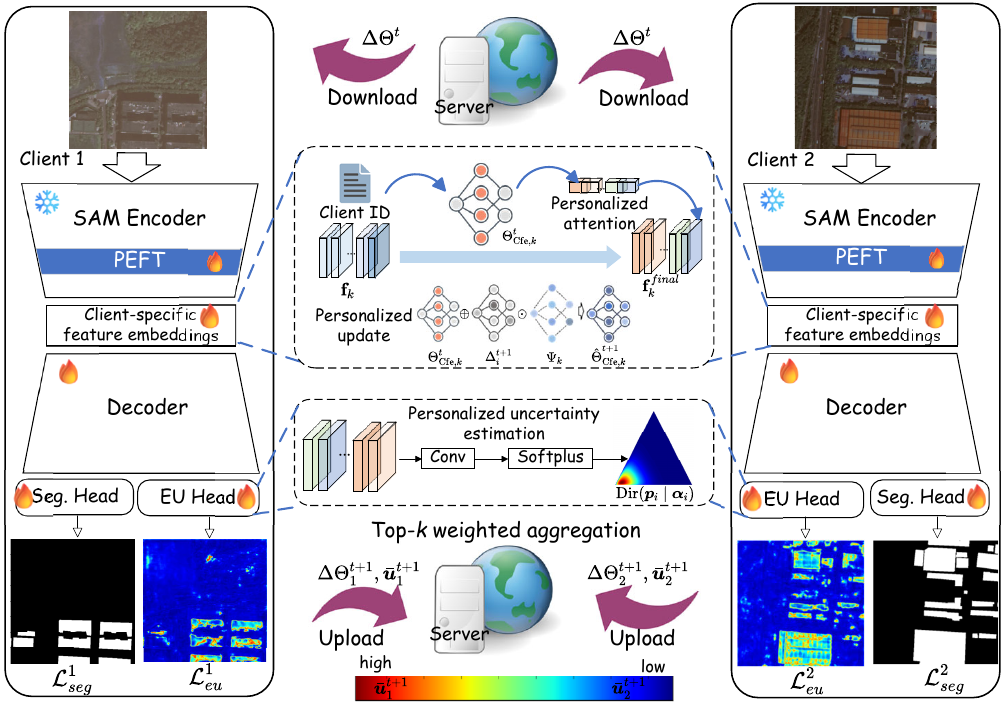}
	\caption{{The framework of FedEU. The local network is a typical encoder-decoder framework with a segmentation head and an evidential uncertainty (EU) head. Furthermore, the CFE module is introduced to enable each client to assign personalized attention to different channels and calibrate its feature representations accordingly through dynamic parameter update.  In the local training, the local model generates predictive evidential uncertainties via the EU head. In the global aggregation, the Top-$k$ uncertainty-based weights are used to guide model parameter aggregation on the server side.}}
	\label{structure}
\end{figure*}

\subsection{Federated Image Segmentation}
RSIS has evolved by exploiting architectures such as U-Net, Transformers, and CNN-Transformer hybrids to enhance representation learning and semantic understanding capabilities \cite{10258445,10458980,10298027,10026298} based on centralized training. 
On this basis, large-scale pre-trained foundation models can be effectively adapted to downstream RSIS tasks through fine-tuning with only a limited amount of labeled data, thereby achieving strong generalization performance \cite{10637992,10485462}.
Meanwhile, FL has emerged as a new optimization paradigm applicable to diverse models and tasks \cite{li2023fedfusion,li2024feddiff, jia2024adaptive}.
Specifically, the classical FedAvg method \cite{mcmahan2017communication} aggregated local model updates by averaging to train a global model, but suffers from performance drops under non-IID data. 
To alleviate this, approaches such as FedProx \cite{li2020federated} and FedProto \cite{tan2022fedproto} aim to mitigate client drift and enhance model aggregation across local clients by guiding local models toward the global optimum.
To address spatial structure and fine-grained feature alignment in the pixel-wise segmentation tasks, FedSeg \cite{miao2023fedseg} proposed a modified cross-entropy loss and pixel-level contrastive learning to align local updates with the global model. Furthermore, FedUKD \cite{kanagavelu2023fedukd} integrated knowledge distillation into a federated U-Net model for land use classification.
Federated RSIS remains under-explored and faces the challenges of high spectral and spatial heterogeneity inherent in remote sensing data, with the increased complexity of model optimization in federated environments.

\subsection{Federated Fine-Tuning}
Recently, parameter-efficient fine-tuning of pretrained models for RSIS has drawn sustained attention  \cite{10637992,11063320} as it reduces the cost of model adaptation by updating only a small subset of parameters. 
Federated fine-tuning combines the power of pre-trained large models with the flexibility of federated learning to adapt these models to specific, decentralized datasets while reducing the communication overheads \cite{chen2024feddat,saha2025fedpia}.
By replacing traditional federated model training with PEFT, participants are allowed to collaboratively optimize prompts instead of full model weights \cite{10210127,10095356,deng2024unlocking}. 
Recent FedFMS \cite{Liu_FedFMS_MICCAI2024} and FLAP-SAM frameworks \cite{10.1007/978-3-031-77610-6_21} proposed parameter-efficient approaches to adapt the Segment Anything Model (SAM) within federated frameworks, balancing its powerful representations with federated learning constraints while achieving strong performance on distributed datasets.
The straightforward combination of centralized PEFT algorithms with standard FL methods struggles to align locally fine-tuned models with global objectives, resulting in inconsistent segmentation accuracy across diverse domains. Moreover, the high uncertainty and bias in local updates can compromise the reliability of global aggregation, increasing the risk of overfitting to local distributions and degrading overall generalization.

\subsection{Personalization and Regularization in Federated Optimization}
Recent advances in FL have introduced various personalization and regularization strategies across clients while retaining the shared knowledge \cite{tan2022towards,sabah2024model}. These approaches can be broadly categorized into two categories: local calibration and dual-regularization. 
The local calibration approach attempts to dynamically adjust client updates based on local data characteristics and mitigate inter-client data heterogeneity \cite{wang2022personalizing,zhang2023fedala}. For example, Per-FedAvg \cite{NEURIPS2020_24389bfe} utilized meta-functions to initialize the local models. FedALA \cite{zhang2023fedala} proposed adaptive local aggregation that dynamically adjusts client updates based on local data characteristics.  Moreover, customized model architecture searching and model structure pruning have been exploited to achieve the specificity of local models \cite{10858749,hu2025heterogeneity,10496080}.
For the dual-regularization approach, global consistency and local adaptation are jointly constrained in the training process \cite{zhang2023federated,10194959}. For instance, in FedTGP \cite{zhang2024fedtgp} and FedProto \cite{tan2022fedproto}, the derived deviations across local and global prototypes were utilized to optimize global model updates via an attention-weighted aggregation scheme. And in turn, the global prototypes were used to constrain the local model training. Global geometric priors can be used to guide sample generation and simulate cross-domain distributions \cite{ma2025geometric}. More recently, pFedCSPC \cite{10605121} enhanced client collaboration by providing personalized initial models through adaptive aggregation and leveraging global prototypes to guide local representation learning, thereby alleviating data imbalance and overfitting.
While existing methods can alleviate bias from client data heterogeneity, they neglect the impact of model-specific epistemic variations during global aggregation, especially the unreliable predictions. In comparison, we propose an uncertainty-aware strategy that identifies high-risk predictions, calibrates local features, and refines global aggregation to better align with client-specific objectives.
\section{Methodology}\label{sec3}
As shown in Fig.~\ref{structure}, the proposed FedEU is built on an end-to-end decoder-based RSIS model, where the encoder is adapted from SAM with a PEFT module for model tuning. Moreover, it incorporates client-specific embeddings and evidential uncertainty estimation in local training.
These uncertainties are further used to guide model parameter aggregation on the server side through a TUW method. Two prediction heads are equipped after the decoder, one for segmentation outputs and the other for evidential uncertainty modeling.

\subsection{Preliminaries}
In a typical FL framework, a centralized server is responsible for orchestrating the distributed training process among \( K \) participating clients, indexed by \( \mathcal{K} = \{1, 2, \dots, K\} \). Each client \( k \) possesses its own private dataset, denoted as $\mathcal{D}_k = \{(x_{k,j}, y_{k,j})\}_{j=1}^{N_k}$, where \( x_{k,j} \) and \( y_{k,j} \) represent the \( j \)-th input sample and corresponding label from the local dataset of client \( k \), respectively.

The goal of FL is to collaboratively train a unified global model with identical architecture and hyperparameter settings across clients, without exchanging raw data. The optimization process seeks to minimize a weighted global loss function, formulated as:  
\begin{equation}
\min_{\Theta} J(\Theta) = \sum_{k \in \mathcal{K}} q_k \mathcal{L}^k(\Theta; \mathcal{D}_k),
\label{eq:global_obj}
\end{equation}  
where \( \Theta \) represents the global model parameters, \( \mathcal{L}^k(\cdot; \cdot) \) indicates the local loss function evaluated on client \( k \)'s data, and \( q_k \) is a non-negative weight assigned to each client, satisfying \( \sum_{k \in \mathcal{K}} q_k = 1 \). Typically, \( q_k \) is determined proportionally to the local data volume as \( q_k = \frac{N_k}{N} \), where the total number of training samples is \( N = \sum_{k \in \mathcal{K}} N_k \).

In FedEU, we aim to collaboratively learn individual local models $\{{\Theta}_1, \ldots, {\Theta}_K\}$ for each client, without exchanging the private data. The overall segmentation network $\Theta$ can be divided into several components, including the frozen encoder $\Theta_{\text{En}}$, the trainable PEFT module $\Theta_{\text{P}}$, the CFE module $\Theta_{\text{Cfe}}$, the decoder $\Theta_{\text{De}}$, the segmentation head $\Theta_{\text{Seg}}$, and the EU head $\Theta_{\text{Dir}}$.

\subsection{Client-Specific Feature Embeddings}
\begin{figure}[htp]
	\centering
\includegraphics[width=\linewidth]{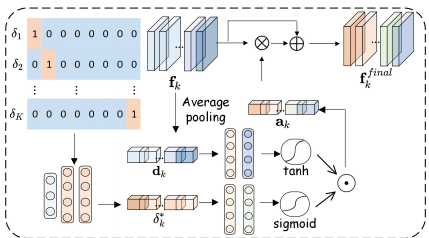}
	\caption{Illustration of the CFE module, which is inspired by \cite{wang2022personalizing}. It employs a personalized channel attention mechanism by embedding domain identifiers into the feature representations and three MLP layers are adopted to perform gated channel attention. The layer parameters are updated via an element-wise strategy considering the dynamics of the training process.}
	\label{cfe}
\end{figure}

The client embedding mechanism aims to encode client-specific characteristics, such as feature variations in data distributions, to construct effective model personalization.
Considering the underlying distributional variance of image data across clients, it is desirable for each client to focus on different feature channels \cite{wang2022personalizing}. To this end, the CFE module is introduced based on the channel attention mechanism to guide each client toward personalized feature representation learning along distinct directions, as shown in Fig.~\ref{cfe}.
Specifically, for the \( k \)-th client, the encoded feature of an image at the \( l \)-th encoding stage is denoted as \( \mathbf{f}_k \in \mathbb{R}^{C \times \frac{H}{2^l} \times \frac{W}{2^l}} \), where \( C \) is the channel number, and \( H \) and \( W \) are the input image height and width, respectively. 
To simplify the communication cost, the client embedding \( \delta_k \in \mathbb{R}^K \) was initialized as a one-hot vector, with the length set as the number of clients \( K \), where the \( k \)-th value is $1$ and others are $0$. On this basis, the feature \( \mathbf{f}_k \) is integrated with the client embedding \( \delta_k \) to enhance the client-specific representation. 
To achieve this, the length of  \( \delta_k \)  was first extended to match the dimensionality of \( \mathbf{f}_k \) for improved training stability. Therefore, the updated client embedding \( \delta_k^* \) can be obtained by:
\begin{equation}
\delta_k^* = \mathrm{MLP}_2\big( \mathrm{IN}\big( \mathrm{MLP}_1(\delta_k) \big) \big), 
\end{equation}
where MLP denotes a multi-layer perceptron consisting of a fully connected layer followed by a ReLU activation function, and IN denotes the instance normalization operation.
Next, the global averaging on each channel of the feature \( \mathbf{f}_k \) was performed to generate a channel descriptor that can represent the abundant contextual information while saving computational cost. The global average pooling on the \( l \)-th encoded feature \( \mathbf{f}_k \) is expressed as:
$\mathbf{d}_k = \frac{1}{H_l \cdot W_l} \sum_{h=1}^{H_l} \sum_{w=1}^{W_l} \mathbf{f}^{h, w}_k$, where \( H_l = \frac{H}{2^l} \) and \( W_l = \frac{W}{2^l} \) are the spatial dimensions after \( l \)-th encoding, and \( \mathbf{f}^{h, w}_k \) is the value of the encoded feature at location \( [h, w] \).


To enhance the adaptability of channel attention to client-specific characteristics, a gating-based mechanism is adopted to generate more expressive attention weights. Specifically, the personalized client embedding $\delta_k^*$ and the channel-wise global descriptor $\mathbf{d}_k$ are first transformed through two independent MLP layers. The output of the client embedding branch is activated by the Sigmoid function to generate a gating vector, while the output of the descriptor branch is activated by the Tanh function to retain both positive and negative contextual cues. The final channel attention weights $\mathbf{a}_k$ are obtained by element-wise multiplication of the two branches:
\begin{equation}
\mathbf{a}_k = \sigma\big(\mathrm{MLP}_1(\delta_k^*)\big) \odot \tanh\big(\mathrm{MLP}_2(\mathbf{d}_k)\big),
\end{equation}
\begin{equation}
\mathbf{f}^{{final}}_{k} = \mathbf{f}_k+\mathbf{f}_k\otimes \mathbf{a}_k,
\end{equation}
where $\sigma$ means the Sigmoid function and
$\odot$ denotes the element-wise multiplication. This gating design allows the model to adaptively emphasize or suppress different channels based on both client-specific characteristics and image content, thereby improving the representation capacity for downstream tasks.
Finally, $\mathbf{f}^{{final}}_{k}$ is fed into the decoder used for further processing in the model.


\subsection{Uncertainty-Guided Local Training}
Evidential deep learning focuses on quantifying and managing the epistemic uncertainty in model predictions \cite{sensoy2018evidential}. 
In the context of local training, this approach enables each client to train models on their own data while incorporating uncertainty measures that reflect the reliability of the learned parameters. By using evidential learning, the local model in FedEU not only provides predictions but also estimates the uncertainty associated with those predictions to enhance model robustness.

For the $i$-th sample, the evidence $e_{c i}$ derived for the $c$-th class can be used to quantify the support provided by data for assigning a sample to a specific class. Then,  the variable $\alpha_{c i}=e_{c i}+1$ can be used to parameterize the Dirichlet distribution. 
By placing the Dirichlet distribution over the prior 
$\boldsymbol{p}_i=\left[p_{1 i}, \ldots, p_{{C} i}\right]$, the distribution can be defined as follows
\begin{equation} 
\mathrm{Dir}\left(\boldsymbol{p}_i \mid \boldsymbol{\alpha}_i\right)=\frac{1}{B\left(\boldsymbol{\alpha}_i\right)} \prod_{c=1}^{{C}} p_{c i}^{\alpha_{c i}-1},
\label{eq:dir}
\end{equation}
where ${{C}}$ represents the number of classes in a classification problem, where $\boldsymbol{\alpha}_i=\left[\alpha_{1 i}, \ldots, \alpha_{{C} i}\right]$ and $B\left(\boldsymbol{\alpha}_i\right)$ are multinomial Beta functions. Then, the uncertainty can be obtained as inversely proportional to the total evidence:
\begin{equation}  
\mathbf{U}_i =C/\sum_{c=1}^{C}(e_{ci}+1).
\label{eq:uncertainty}
\end{equation}


Subsequently, a loss function is formulated to minimize the Bayesian risk associated with the squared error loss, expressed as \( \left\|\boldsymbol{y}_i - \boldsymbol{p}_i \right\|_2^2 \). In addition, a Kullback-Leibler (KL) divergence term is incorporated as a regularizer to discourage excessive uncertainty in the model's predictions:
\begin{equation}
\begin{aligned}
\mathcal{L}^{k}_{eu} &= \sum_{i=1}^{N_k} \int \left\|\boldsymbol{y}_i-\boldsymbol{p}_i\right\|_2^2 \frac{1}{B\left(\boldsymbol{\alpha}_i\right)} \prod_{c=1}^{C} \boldsymbol{p}_{c i}^{\alpha_{c i}-1} d \boldsymbol{p}_i \\
 &+ \lambda \sum_{i=1}^{N_k} KL\left[D\left(\boldsymbol{p}_i \mid \tilde{\boldsymbol{\alpha}}_i\right) \| D\left(\boldsymbol{p}_i \mid \langle 1, \ldots, 1\rangle\right)\right]
\end{aligned}
\end{equation}
where \( D\left(\boldsymbol{p}_i \mid \langle 1, \ldots, 1 \rangle \right) \) refers to a uniform Dirichlet distribution, where the vector \( \langle 1, \ldots, 1 \rangle \) implies that each class shares an identical concentration parameter, corresponding to a state of maximum uncertainty.

Therefore, the overall training objective for each client is the combination of the binary cross-entropy-based segmentation loss $\mathcal{L}_{seg}$  and the evidential uncertainty loss $\mathcal{L}_{eu}$, as follows:
\begin{equation}
\mathcal{L}^{k} ( \Theta ) = \mathcal{L}_{seg} (\Theta \setminus \Theta_{\text{Dir}}\cup\Theta_{\text{En}}) +\mu\mathcal{L}_{eu} (\Theta \setminus \Theta_{\text{Seg}}\cup\Theta_{\text{En}}),
\end{equation}
where the hyperparameter $\mu$ controls the contribution of the two terms.

Following this, each client \( k \) updates its local model parameters \( \Theta_k^t \) via mini-batch stochastic gradient descent (SGD) to progressively approach optimal values:
\begin{equation}
\Theta_k^{t+1} = \Theta_k^t - \eta \nabla \mathcal{L}^k \big( \Theta^t \big),
\end{equation}
where \( \eta \) denotes the learning rate. The resulting updated parameters \( \Theta_k^{t+1} \) are then used both for participation in global aggregation during communication round \( t \), and as the starting point for local training in round \( t + 1 \).

 \subsection{{Uncertainty-Guided  Global Aggregation}}
In FedEU, the TUW strategy adopts an uncertainty-constrained model aggregation by prioritizing the reduction of risks in the most uncertain and vulnerable regions, rather than simply minimizing the average risk.
According to information theory, high-uncertainty regions are typically more information-dense and harder to compress, making them more representative of the model’s generalization ability. Clients with more uncertain regions are considered less reliable and should have lower aggregation weights to avoid introducing misleading patterns into the global model. Therefore, FedEU enhances robustness by reducing the impact of pseudo-optimal regions and guiding the global model to focus on critical, yet under-learned, knowledge.

To capture the most informative uncertainty regions during each local training round, we adopt a pixel-level Top-$k$ selection strategy for estimating client-level uncertainty. This strategy ensures that clients with more confident local predictions are assigned higher contributions during global model aggregation.
Specifically, for client $k$, we first compute the pixel-wise uncertainty map $\mathbf{U}_{k}[h, w]$ across all local samples. Then, we select the top $\tau$ pixels with the highest uncertainty values, denoted as $\mathcal{R}_c^{\text{Top-}\tau}$, and calculate the average uncertainty over these pixels, as follows:
\begin{equation} 
\bar{\boldsymbol{u}}_{k,i} = \frac{1}{\tau} \sum_{(h, w) \in \mathcal{R}_{k,i}^{\text{Top-}\tau}} \mathbf{U}_{k,i}[h, w].
\end{equation} 
Then, the average $\bar{\boldsymbol{u}}_k^t$ of all samples in this training round is computed for the client, which can be obtained by 
$\bar{\boldsymbol{u}}_k^t = \frac{1}{N_k} \sum_{i=1}^{N_k}  \boldsymbol{\bar u}_{k,i}^t$.
Then, the confident deviations across all clients can be used to quantify the federated weights $\boldsymbol{w}^t_k$ of the $k$-th client as follows:
\begin{equation} 
\boldsymbol{\omega}^{t}_k= \begin{cases}N_k / \sum_{k^{\prime} \in \mathcal{K}^t} N_{k^{\prime}}, & t=0 \\ (1-\bar{\boldsymbol{u}}^{t}_k) / \sum_{k^{\prime} \in \mathcal{K}^t} (1-\bar{\boldsymbol{u}}^{t}_{k^{\prime}}). & t \geq 1\end{cases}\label{eq:weight}
\end{equation} 
Here, clients with higher $\boldsymbol{\omega}^{t}_k$ will contribute more to the global aggregation, ensuring that well-trained and more confident clients have a greater influence in the federated optimization process. This dynamic reweighting mechanism effectively mitigates the impact of uncertain clients and enhances overall model performance. 



At the end of each communication round, the global model is updated by aggregating the local parameter changes using an adaptive weighting mechanism, as defined below:
\begin{equation} 
\Delta \Theta^{t+1} = \sum_{k \in \mathcal{K}^t} \boldsymbol{\omega}_k^t \odot \Delta \Theta_k^{t+1}, \quad \forall k \in \mathcal{K}^t
\end{equation}
where \( \boldsymbol{\omega}_k^t \) represents the adaptive weight that reflects the contribution of client \( k \)'s local model to the global decision boundary. The aggregated update \( \Delta \Theta^{t+1} \) is then applied to refine the global model parameters: $\Theta^{t+1} = \Theta^t + \Delta \Theta^{t+1}.$
Each local update from client \( k \) is computed as the difference between two successive local model states:
$\Delta \Theta_k^{t+1} = \Theta_k^{t+1} - \Theta_k^t$.

\subsection{Adaptive Update of CFE}
In the above-mentioned global aggregation process, all parameters of a client model share the same weight.
To incorporate fine-grained personalization into the local model, the update of the CFE module is performed via an adaptive element-wise aggregation manner as follows:
\begin{equation}
\begin{aligned}
\hat{\Theta}^{t+1}_{\mathrm{Cfe},k} &:= \Theta^{t}_{\mathrm{Cfe},k} + (\Theta^{t+1}_{\mathrm{Cfe}} - \Theta^{t}_{\mathrm{Cfe},k}) \odot \Psi_k,
\\ s.t. \quad & \xi(\psi) = \max(0, \min(1, \psi)), \quad \forall \psi \in \Psi_k
\end{aligned}\label{eq:ala}
\end{equation}
where the parameters of CFE, $\hat{\Theta}^{t+1}_{\mathrm{Cfe},k}$, are updated through an adaptive fusion of the global model and local model updates, guided by the coefficient matrix $\Psi_k$. This process can be regarded as the model calibration after each client performs local model training \cite{zhang2023fedala}. The clipping operation $\xi( \cdot)$ is applied for regularization to make sure $\psi$ falls into $[0,1]$. The coefficient matrix $\Psi_k$ can be learned by the gradient-based optimization method:
\begin{equation}
\Psi_k \leftarrow \Psi_k - \eta \nabla_{\Psi} \mathcal{L}^{k}(\hat{\Theta}_{\mathrm{Cfe},k}^{t}, {\mathcal{D}}_{k}^{t} ; \Theta^{t}_{\mathrm{Cfe}}).\label{weight_update}
\end{equation}
During this process, all trainable parameters of the segmentation network are frozen, except those in the CFE module.


\begin{algorithm}[htb]
\caption{FedEU}\label{algorithm_fedpm}
\KwIn{$\Theta$: model parameters; $T$: number of communication rounds; $E$: number of local training epochs; $\eta$: learning rate; $\mu$: evidential loss ratio; $\{\mathcal{C}_k\}_{k=1}^K$: all clients; $\mathcal{S}$: the server}
\KwOut{Personalized local model parameters $\{{\Theta}_1, \ldots, {\Theta}_K\}$}

Initialize the global model $\Theta$\;
\For{$t = 0, \dots, T-1$}{
    Select training clients $\mathcal C_{k} \in \{\mathcal C_{k}\}_{k=1}^{\mathcal K^t}$\;
    $\mathcal C_{k}$ downloads $\Theta^{t}$ from $\mathcal S$\;
    \If{$t > 0$}{
        $\Theta_k^t \leftarrow \Theta^t \cup \Theta_{\text{Dir},k}^t$\;
    }
    \tcc{Local Training}
    \For{each client}{
        $\Theta_{k}^{t+1} \leftarrow \Theta_{k}^{t} - \eta \nabla J_{k}(\Theta^{t})$\;
        $\mathcal C_{k}$ computes $\mathbf{U}^{t}_k$ by \eqref{eq:uncertainty}\;
        $\Delta\Theta_{k}^{t+1} \leftarrow \Theta_{k}^{t+1} - \Theta_{k}^{t}$\;
        $\mathcal C_{k}$ uploads ${\boldsymbol{\bar u}}_k^t$ and $\Delta\Theta_{k}^{t+1}$ to $\mathcal S$\;
    }
    \tcc{Global Aggregation}
    Calculate the weight $w^{t}_k$ by \eqref{eq:weight}\;
    $\Delta\Theta^{t+1} \leftarrow \sum {w}_{k}^{t} \odot \Delta\Theta_{k}^{t+1}$\;
    $\Theta^{t+1} \leftarrow \Theta^{t} - \Delta\Theta^{t+1}$\;
    \tcc{Client Embeddings}
    \For{each client}{
        Download $\Theta^{t+1}$ from $\mathcal{S}$\;
        Update $\Psi_k$ by \eqref{weight_update}\;
        Personalize the CFE by \eqref{eq:ala}\;
    }
}
\end{algorithm}

\begin{table}[htp]
	\centering	
	\caption{Description of the experimental setups about clients and samples with an image size of 512$\times$512 .}
	\label{table_dataset}
    \setlength{\tabcolsep}{5pt} 
	\begin{tabular}{cccc}\toprule
		Dataset      & Clients & Fine-tuning samples  & Test samples  \\ \hline
		{GF-7}              &{6}         &{75 / 43 / 43 / 44 / 45 / 58}     &{101}    \\ 
		{GLH}              &{5}         &{30 / 30 / 30 / 30 / 30 / 30}     &{250}  \\  
		{GVLM}              &{6}         &{71 / 93 / 141 / 81 / 161 / 72}     &{208}   \\    \toprule
	\end{tabular}
\end{table}

\begin{figure*}[htp]
	\centering
	\includegraphics[width=\linewidth]{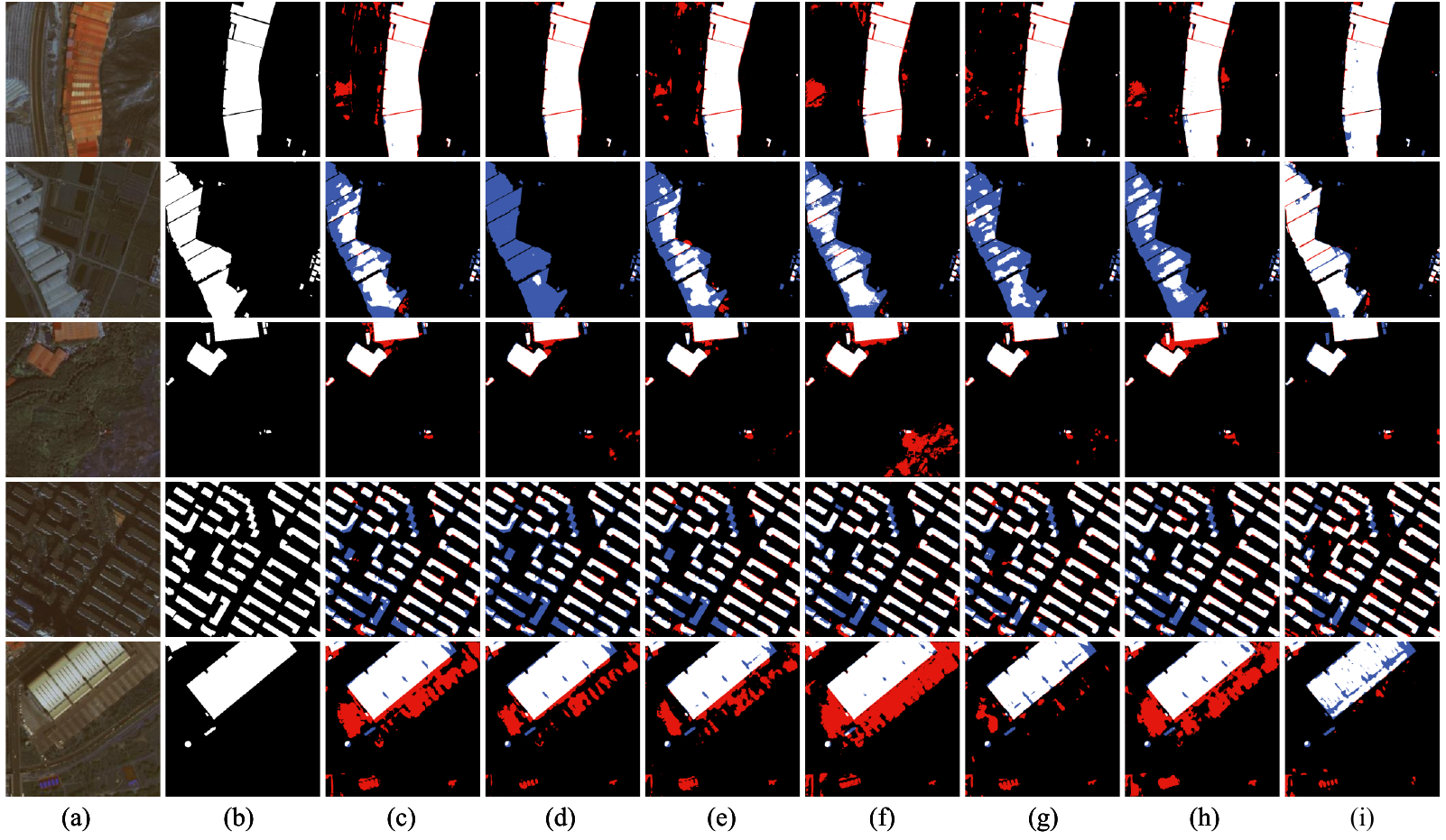}
      \vspace{-8pt}
	\caption{Visual comparison of building extraction results from various FL algorithms on the GF-7 dataset. (a) Original image. (b) Ground truth. (c)-(h) The segmentation results obtained by (c) FedAvg, (d) FedProx, (e) FedSeg, (f) FedTGP, (g) Per-FedAvg, (h) FedBN and (i) FedEU. True positives, true negatives, false positives and false negatives are denoted by white, black,  red, and blue, respectively.}
	\label{fig_result1}
\end{figure*}

\begin{table*}[htbp]
\centering
\caption{The test accuracies (\%) on the GF-7 dataset.}
\label{table_result1}
\setlength{\tabcolsep}{7pt}
\begin{tabularx}{\textwidth}{l*{14}{c}}
\toprule
\multirow{2}{*}{Method} & \multicolumn{2}{c}{Chongqing} & \multicolumn{2}{c}{Guangzhou} & \multicolumn{2}{c}{Lanzhou} & \multicolumn{2}{c}{Ningbo} & \multicolumn{2}{c}{Shenzhen} & \multicolumn{2}{c}{Tianjin} & \multicolumn{2}{c}{Total} \\
& IoU & OA & IoU & OA & IoU & OA & IoU & OA & IoU & OA & IoU & OA & IoU & OA \\
\hline
FedAvg \cite{mcmahan2017communication}  & 45.39 & 89.37 & 46.46 & 88.97 & 47.00 & 84.67 & 50.67 & 87.19 & 46.16 & 86.87 & 56.19 & 90.42 & 48.66 & 88.18 \\
FedProx \cite{li2020federated}  & 55.27 & 93.17 & 41.67 & 88.52 & 42.71 & 85.97 & 55.84 & 90.07 & 45.33 & 87.30 & 52.63 & 90.32 & 49.73 & 89.68 \\
FedSeg \cite{miao2023fedseg} & 52.19 & 90.49 & 49.54 & 90.00 & 57.40 & 91.31 & 52.86 & 89.24 & 48.21 & 88.67 & 52.49 & 90.32 & 52.11 & 90.49 \\
FedTGP \cite{zhang2024fedtgp}  & 52.03 & 92.09 & 47.83 & 89.66 & 49.21 & 87.04 & 56.67 & 90.33 & 51.22 & 89.03 & 55.77 & 91.02 & 52.29 & 90.72 \\
Per-FedAvg \cite{NEURIPS2020_24389bfe}  & 50.90 & 93.79 & 56.22 & 90.90 & 49.80 & 91.35 & 54.07 & 89.34 & 45.35 & 90.26 & 54.09 & 89.35 & 51.74 & 91.06 \\
FedBN \cite{li2021fedbn} & 52.14 & 92.16 & 47.08 & 89.43 & 53.28 & 89.09 & 54.93 & 89.77 & 49.31 & 88.38 & 52.00 & 90.45 & 51.46 & 89.88 \\
FedEU (Ours) & \textbf{57.66} & \textbf{95.51} & \textbf{61.09} & \textbf{93.53} & \textbf{61.28} & \textbf{92.52} & \textbf{65.23} & \textbf{92.38} & \textbf{53.57} & \textbf{90.38} & \textbf{61.15} & \textbf{93.80} & \textbf{55.32} & \textbf{91.26} \\
\bottomrule
\end{tabularx}
\end{table*}



\subsection{Implementation}
The implementation of the proposed FedEU framework is summarized in Algorithm~\ref{algorithm_fedpm}. At each communication round, the server broadcasts the current global model parameters to the selected clients. Each client initializes its local model using the global model and incorporates the CFE module to adaptively calibrate the representation space. Local training is performed using a mini-batch stochastic gradient descent algorithm. During training, the evidential uncertainty of each sample is estimated via the Dirichlet-based output of the evidential classifier.
Within each client, a Top-$k$ subset of training samples with the highest uncertainty is selected to compute the client-level local uncertainty. 
Upon completion of local training, clients evaluate the uncertainty of their models and upload the model parameters along with the uncertainty estimates to the server. The server then performs an uncertainty-aware aggregation, where local models with lower overall uncertainty receive higher weights in the global model update, reducing the influence of models with high uncertainty or severe distribution drift. The aggregated global model is then broadcast back to clients for the next training round. This iterative process continues until convergence.

The image encoder adopts the ViT-B variant of Segment Anything Model (SAM), pre-trained on the SA-1B dataset \cite{kirillov2023segment}. It contains 12 Transformer blocks, each integrated with a lightweight fine-tuning module (e.g., Adapter). Moreover, the encoder is followed by a normal CNN-based decoder that produces the final segmentation outputs.
To be noticed, our work primarily focuses on federated optimization, while the fine-tuning strategy adopts the widely used Adapter-tuning-based PEFT methods. Specifically, the core components of the adapter consist of a downscale MLP layer, a ReLU activation function and an upscale MLP layer. Specifically, it interleaves transformer layers with a feed-forward bottleneck module, using a skip connection to adapt the layer’s output before passing it to the next layer.

\begin{figure*}[htp]
	\centering
	\includegraphics[width=\linewidth]{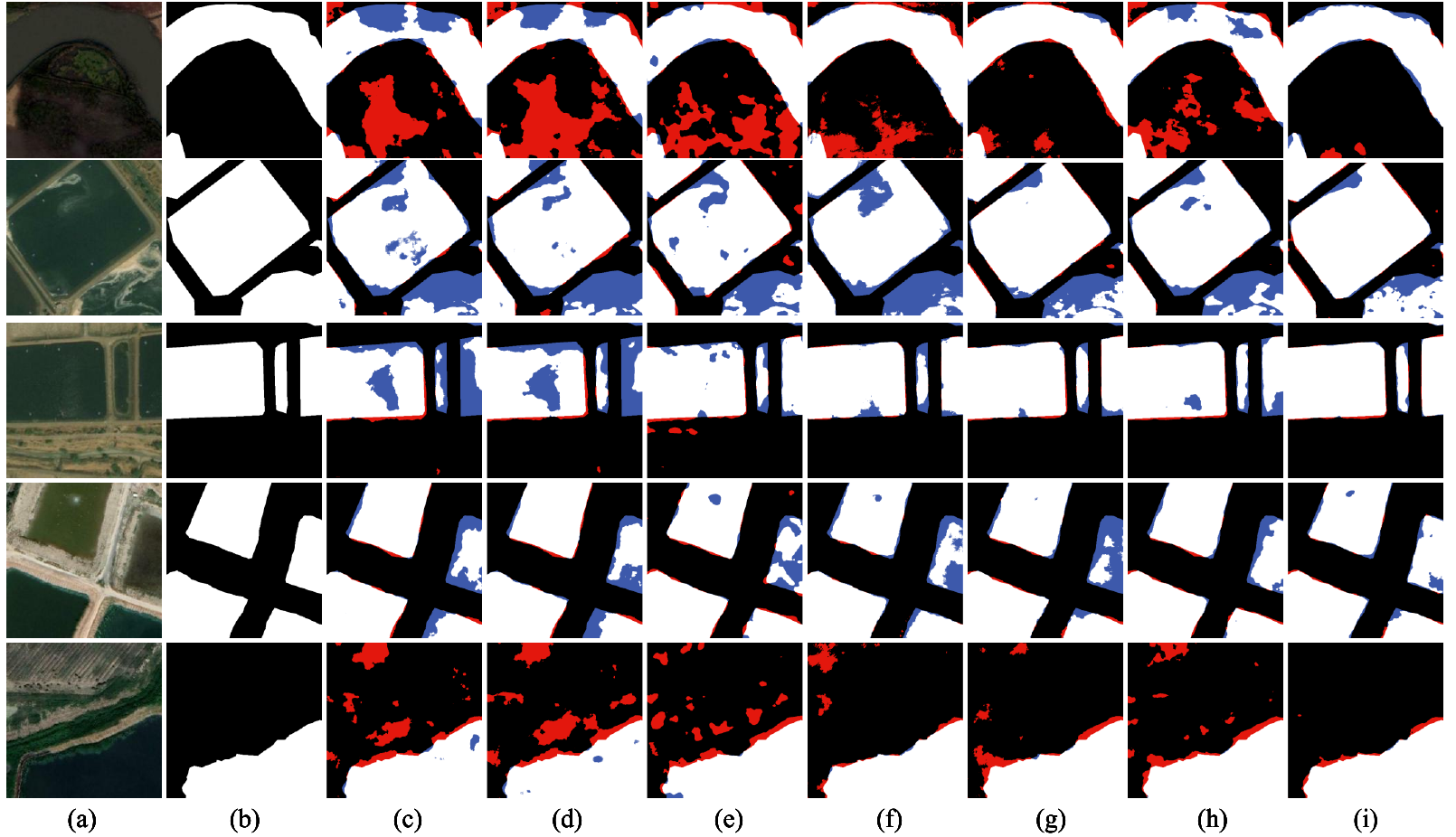}
    \vspace{-8pt}
	\caption{Visual comparison of water extraction results from various FL algorithms on the GLH-water dataset. (a) Original image. (b) Ground truth. (c)-(h) The segmentation results obtained by (c) FedAvg, (d) FedProx, (e) FedSeg, (f) FedTGP, (g) Per-FedAvg, (h) FedBN and (i) FedEU. True positives, true negatives, false positives and false negatives are denoted by white, black,  red, and blue, respectively.}
	\label{fig_result2}
\end{figure*}

\section{Experiments}\label{sec4}
\subsection{Datasets}
To comprehensively evaluate the proposed approach, we conduct experiments on three widely used RSIS datasets, including the GaoFen-7 (GF-7) Building Dataset, the GLH-Water Dataset, and the Global Very-High-Resolution Landslide Mapping (GVLM) Dataset. These datasets cover diverse land-cover types and exhibit notable spectral heterogeneity, making them suitable for federated model learning evaluation. Detailed dataset configurations are summarized in Table~\ref{table_dataset}. We use a relatively small number of samples for fine-tuning the pre-trained models and a larger set for testing.

\subsubsection{GaoFen-7 (GF-7) Building Dataset}
The GF-7 Building Dataset \cite{chen2024benchmark} offers high-resolution imagery at \( 0.65 \) meters, collected from several urban regions in China, including Tianjin, Lanzhou, Chongqing, Ningbo, Guangzhou, and Shenzhen. It contains a total of $5,175$ image-label pairs, each comprising \( 512 \times 512 \) pixel tiles, and spans an area of approximately $573.17$ square kilometers.
The dataset spans various geographical locations, providing a broad range of samples. Each client in the FL setup receives a sub-dataset linked to a specific city, ensuring the non-IID nature required for algorithm evaluation. 

\subsubsection{GLH-Water Dataset}
The GLH-Water Dataset \cite{li2024glh} contains $250$ satellite images and surface water annotations, which are globally distributed and represent various water body types, such as rivers, lakes, ponds in forests, irrigation fields, barren areas, and urban regions. Each original image has dimensions of \( 12{,}800 \times 12{,}800 \) pixels and a ground sampling distance of \( 0.3 \) meters. For experimental purposes, the dataset is partitioned into five subsets according to geographic regions, and the large-sized images are further cropped into smaller patches of size \( 512 \times 512 \) pixels.


\subsubsection{Global Very-High-Resolution Landslide Mapping (GVLM) Dataset}  
The GVLM dataset \cite{zhang2023cross} is a large-scale benchmark for high-resolution landslide detection. It comprises 17 pairs of ultra-fine resolution remote sensing images, each with a spatial resolution of \( 0.59 \) meters. The data is organized into subsets corresponding to six geographically diverse landslide sites—located in Vietnam, China, India, Chile, Kyrgyzstan, and Georgia—each assigned to a distinct client for federated landslide extraction tasks.  Due to pronounced spectral diversity and intensity fluctuations across these regions, this dataset presents considerable challenges for federated learning models to achieve robust and consistent training performance.


\begin{table*}[htp]
\centering
\caption{The test accuracies (\%) on the GLH dataset.}
\label{table_result2}
\setlength{\tabcolsep}{9.3pt}
\begin{tabularx}{\textwidth}{l*{12}{c}}
\toprule
\multirow{2}{*}{Method} & \multicolumn{2}{c}{Client 1} & \multicolumn{2}{c}{Client 2} & \multicolumn{2}{c}{Client 3} & \multicolumn{2}{c}{Client 4} & \multicolumn{2}{c}{Client 5} & \multicolumn{2}{c}{Total} \\
& IoU & OA & IoU & OA & IoU & OA & IoU & OA & IoU & OA & IoU & OA \\
\hline
FedAvg \cite{mcmahan2017communication} & 62.16 & 80.12 & 76.19 & 89.66 & 71.51 & 88.29 & 81.38 & 90.77 & 51.80 & 72.81 & 64.00 & 81.17 \\
FedProx \cite{li2020federated}  & 62.67 & 79.96 & 75.98 & 89.52 & 73.25 & 88.98 & 82.00 & 91.40 & 51.52 & 72.80 & 64.13 & 81.98 \\
FedSeg \cite{miao2023fedseg} & 62.37 & 79.41 & 71.40 & 87.82 & 72.53 & 88.41 & 83.31 & 91.95 & 53.97 & 73.78 & 63.95 & 81.56 \\
FedTGP \cite{zhang2024fedtgp} & 71.74 & 85.98 & 82.48 & 92.86 & 84.14 & 93.87 & 87.74 & 94.02 & 55.70 & 74.39 & 70.22 & 85.34 \\
Per-FedAvg \cite{NEURIPS2020_24389bfe} & \textbf{77.45} & 88.41 & 91.22 & 96.52 & \textbf{93.13} & \textbf{97.42} & 95.69 & 97.99 & 68.19 & 86.42 & 80.40 & 91.11 \\
FedBN \cite{li2021fedbn}  & 69.91 & 84.63 & 78.12 & 90.35 & 81.68 & 92.10 & 85.99 & 93.94 & 46.84 & 68.01 & 72.51 & 85.81 \\
FedEU (Ours) & 77.43 & \textbf{89.06} & \textbf{92.29} & \textbf{97.01} & 92.46 & 97.16 & \textbf{95.81} & \textbf{98.04} & \textbf{70.64} & \textbf{87.12} & \textbf{85.73} & \textbf{93.68} \\
\bottomrule
\end{tabularx}
\end{table*}

\subsection{Implementation Details}
The proposed FedEU is compared with various state-of-the-art FL algorithms, including FedAvg \cite{mcmahan2017communication}, FedProx \cite{li2020federated}, 
FedTGP \cite{zhang2024fedtgp}, FedSeg \cite{miao2023fedseg}, Per-FedAvg \cite{NEURIPS2020_24389bfe} and FedBN \cite{li2021fedbn}, where the last two approaches are personalized optimization methods.

In the process of FedEU, each client adopts the same hyperparameters. The Adam optimizer was used for gradient updates. The number of communication rounds $T$ was set to $200$, and the learning rate and mini-batch size for local training were set to $0.001$ and $8$, respectively. Additionally, the number of local training epochs $E$ is set to $5$. For the hybrid loss, the weight $\mu$ in the total loss is set to $0.8$.



Two widely used metrics, Overall Accuracy (OA) and Intersection over Union (IoU), were employed to evaluate the proposed method in the experiment.
The OA metric evaluates the model’s ability to correctly classify both positive and negative samples. In contrast, IoU quantifies the overlap between predicted and ground truth regions by comparing their intersection to their union. While OA provides a broad evaluation of a model’s accuracy, IoU is particularly valuable for assessing spatial consistency in object segmentation.



\begin{figure*}[htp]
	\centering
	\includegraphics[width=\linewidth]{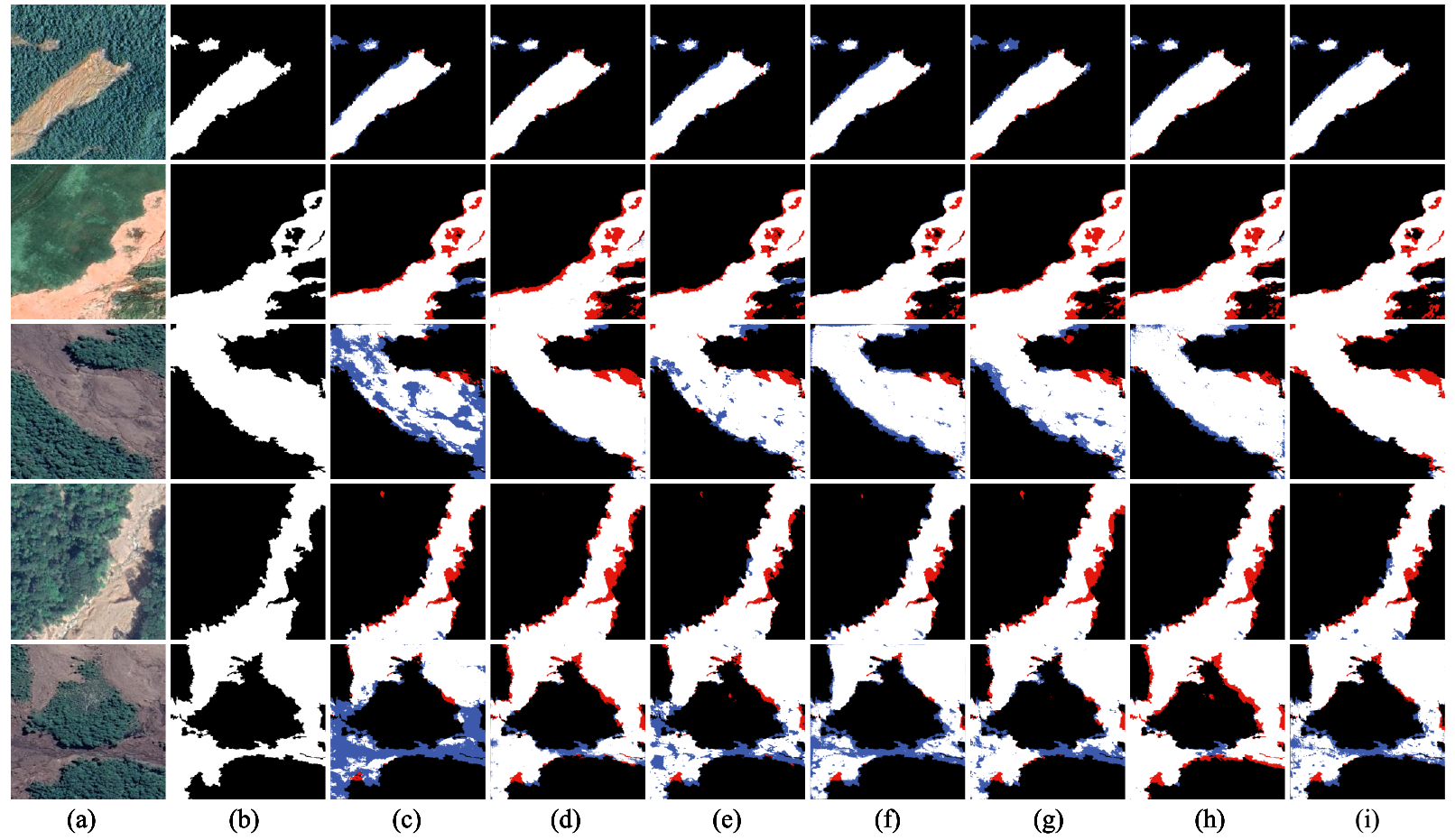}
	  \vspace{-8pt}
      \caption{Visual comparisons on the GVLM dataset. (a) Original image. (b) Ground truth. (c)-(h) The segmentation results obtained by (c) FedAvg, (d) FedProx, (e) FedSeg, (f) FedTGP, (g) Per-FedAvg, (h) FedBN and (i) FedEU. True positives, true negatives, false positives and false negatives are denoted by white, black,  red, and blue, respectively.}
	\label{fig_result3}
\end{figure*}


\begin{table*}[htbp]
\centering
\caption{The test accuracies (\%) on the GVLM dataset.}
\label{table_result3}
\setlength{\tabcolsep}{7pt}
\begin{tabularx}{\textwidth}{l*{14}{c}}
\toprule
\multirow{2}{*}{Method} & \multicolumn{2}{c}{Vietnam} & \multicolumn{2}{c}{China} & \multicolumn{2}{c}{India} & \multicolumn{2}{c}{Chile} & \multicolumn{2}{c}{Kyrgyzstan} & \multicolumn{2}{c}{Georgia} & \multicolumn{2}{c}{Total} \\
& IoU & OA & IoU & OA & IoU & OA & IoU & OA & IoU & OA & IoU & OA & IoU & OA \\
\hline
FedAvg \cite{mcmahan2017communication} & 83.06 & 99.67 & 47.63 & 98.29 & 35.45 & 98.86 & 71.62 & 90.76 & 82.52 & 97.60 & 66.75 & 95.26 & 59.26 & 96.24 \\
FedProx \cite{li2020federated}  & 83.68 & 99.69 & 47.24 & 98.29 & 35.65 & 98.83 & 71.64 & 90.76 & 82.11 & 97.53 & 66.42 & 95.14 & 59.36 & 96.25 \\
FedSeg \cite{miao2023fedseg} & 84.10 & 99.69 & 46.71 & 98.26 & 35.31 & 98.87 & 71.92 & 90.85 & 81.54 & 97.51 & 66.35 & 95.19 & 58.89 & 96.61 \\
FedTGP \cite{zhang2024fedtgp}  & 83.58 & 99.68 & 47.39 & 98.28 & 66.55 & 99.46 & 71.31 & 90.66 & 82.70 & 97.62 & 47.79 & 91.80 & 64.90 & 96.06 \\
Per-FedAvg \cite{NEURIPS2020_24389bfe}  & \textbf{84.64} & \textbf{99.70} & 51.13 & 98.16 & 64.90 & 99.37 & \textbf{93.54} & \textbf{98.12} & \textbf{85.24} & 97.99 & \textbf{73.93} & \textbf{96.49} & 72.45 & 98.13 \\
FedBN \cite{li2021fedbn}  & 82.12 & 99.66 & \textbf{52.26} & 98.33 & 63.21 & 99.39 & 92.87 & 97.95 & 70.17 & 94.79 & 65.82 & 94.39 & 71.07 & 97.42 \\
FedEU (Ours) & 83.21 & 99.65 & 52.16 & \textbf{98.33} & \textbf{70.78} & \textbf{99.56} & 93.30 & 98.07 & 85.11 & \textbf{98.00} & 73.02 & 96.21 & \textbf{76.63} & \textbf{98.25} \\
\bottomrule
\end{tabularx}
\end{table*}

\subsection{Experimental Results}
\subsubsection{GF-7 Building Dataset}
Fig.~\ref{fig_result1} shows the performance of FedEU and other FL methods on the GF-7 dataset for building extraction. 
While FedAvg and FedProx perform reasonably well on small buildings, they struggle with boundary delineation for large structures, leading to significant false positives. FedTGP and FedSeg partially alleviate data heterogeneity but still face difficulties in modeling complex building structures and capturing personalized features within the global model. Due to non-IID data across clients, existing FL methods often exhibit large omissions within building interiors. 
In contrast, FedEU more accurately distinguishes building boundaries and suppresses noise from data heterogeneity. As shown in Table~\ref{table_result1}, client-level performance varies significantly due to distributional differences. Compared methods fail to balance global generalization and client-specific adaptation, whereas FedEU achieves consistent improvements in IoU and OA, with IoU gains ranging from $3.58\%$ to $6.66\%$ over other methods. While some methods perform well on specific clients, their weak generalization limits overall effectiveness, highlighting FedEU’s advantage in accommodating diverse client needs.

\subsubsection{GLH-Water Dataset}
Fig.~\ref{fig_result2} presents the visualization results on the GLH dataset. Due to varying environmental and lighting conditions across clients, significant data heterogeneity exists, posing challenges for FL. For example, in some cases (e.g., the first and fifth rows), river pixels are difficult to distinguish, leading to frequent false detections and omissions. Methods like FedAvg, FedProx, and FedSeg often produce fragmented results with false negatives, while FedTGP improves smoothness but still suffers from client drift caused by distributional differences.
In contrast, FedEU better handles such heterogeneity, reducing noisy predictions and improving the consistency of water body extraction. As shown in Table~\ref{table_result2}, FedEU outperforms other methods, achieving $5.33\%–21.78\%$ gains in IoU and $2.57\%–12.51\%$ improvements in OA, demonstrating its generalization capability in non-IID scenarios. Specifically, FedEU significantly outperforms other methods on Client 2, Client 4, and Client 5, while Per-FedAvg and FedBN achieve slightly better results on Client 3, respectively. Despite this, the consistently superior average results confirm the robustness and effectiveness of FedEU.

\subsubsection{GVLM Dataset}
As shown in Fig.~\ref{fig_result3}, the heterogeneity of landslide images across clients is more pronounced than that of buildings or water bodies, due to the irregular and fragmented boundaries of landslide regions. These characteristics make accurate segmentation more difficult. Additionally, discontinuities in landslide regions within some clients can propagate through global aggregation, leading to missed detections in others (e.g., rows 3 and 5). FedAvg produces overly consistent boundary predictions across clients, failing to adapt to diverse ground truths. While FedBN, FedTGP, FedSeg, and Per-FedAvg mitigate client drift to some extent using regularization and prototype alignment, they still struggle with precise boundary delineation.
In contrast, FedEU leverages evidential uncertainty to guide more personalized local learning, resulting in better handling of image heterogeneity. As shown in Table~\ref{table_result3}, FedEU achieves the best overall IoU of $76.63\%$ and OA of $98.25\%$, significantly outperforming all competing methods in the aggregated results. While Per-FedAvg achieves slightly better results on the Vietnam, Chile, and Georgia subsets, the consistently strong performance of FedEU across diverse clients highlights its robustness and effectiveness in handling distributional heterogeneity for cross-region RSIS.

\begin{table}[htp]
\centering
\setlength{\tabcolsep}{5pt} 
\caption{Ablation studies (\%) of key components in FedEU on the GF-7.}
\label{ablation}
\begin{tabular}{c c c c c c c c c}
\toprule
CFE & TUW & CQ & GZ & LZ & NB & SZ & TJ & Total \\ 
\hline
    &      & 55.46 & 57.92 & 59.81 & 55.67 & 50.69 & 55.85 & 52.46 \\
\checkmark    &     & 54.45 & 60.65 & 60.33 & 55.87 & 51.78 & 56.80 & 53.28 \\
    & \checkmark    & 54.19 & 59.36 & 60.44 & \textbf{56.25} & 51.70 & \textbf{56.82} & 54.07 \\
\checkmark    & \checkmark    & \textbf{57.62} & \textbf{61.09} & \textbf{61.50} & 55.91 & \textbf{52.75} & 55.15 & \textbf{55.32} \\
\bottomrule  
\end{tabular}
\end{table}

\begin{figure}[htp]
	\centering
\includegraphics[width=1\linewidth]{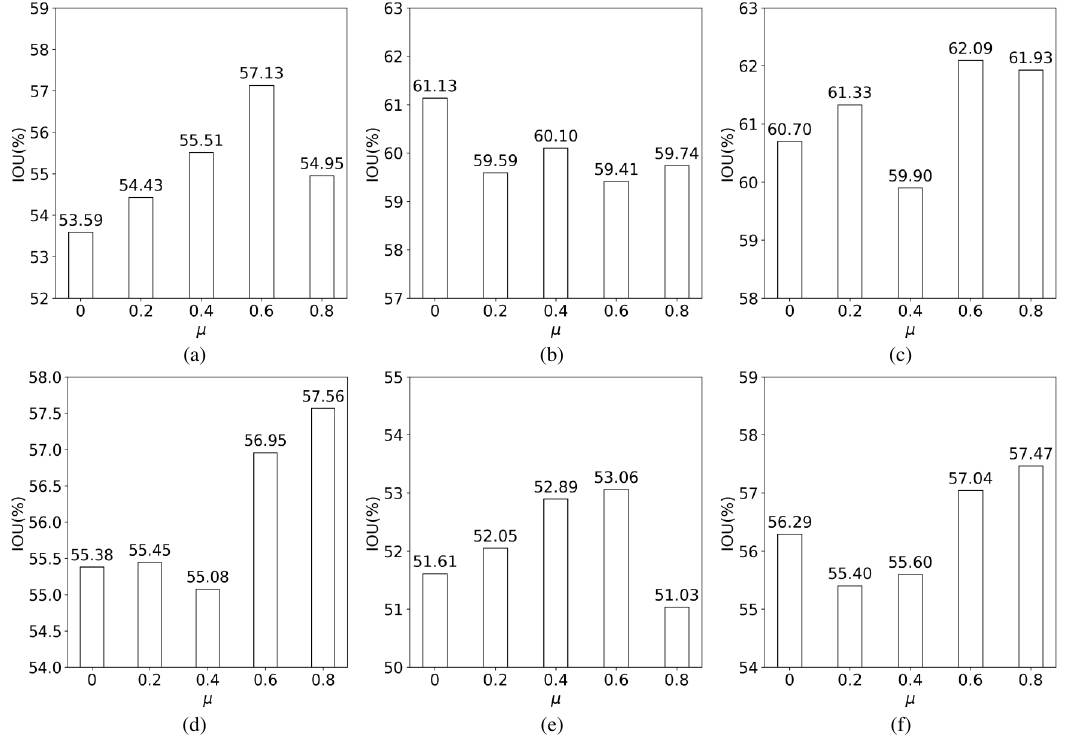}
	\caption{{Analysis of the weight $\mu$ in the EU head on the prediction performance. (a) to (f) are results in the clients of Chongqing, Guangzhou, Lanzhou, Ningbo, Shenzhen and Tianjin, respectively.}}
\label{para_analysis}
\end{figure}

\subsection{Ablation Studies}
To evaluate the contribution of each component in the proposed framework, we conducted an extensive ablation study, with the results summarized in Table~\ref{ablation}, where CQ, GZ, LZ, NB, SZ, and TJ refer to Chongqing, Guangzhou, Lanzhou, Ningbo, Suzhou, and Tianjin, respectively. The results demonstrate that both the CFE module and the uncertainty constraints significantly enhance overall performance. Specifically, removing the CFE module led to degraded performance in areas with pronounced client-specific characteristics, such as Chongqing and Guangzhou, emphasizing the necessity of integrating client-specific information to better capture localized patterns. Moreover, when the TUW and uncertainty constraints for global aggregation were excluded, performance in some cities, such as Lanzhou, slightly improved, while others, including Chongqing and Guangzhou, declined. It showed that the model exhibited diminished personalization across clients due to increased client drift, resulting in a noticeable drop in IoU accuracy. This highlights the critical role of uncertainty-aware aggregation in mitigating the effects of non-IID data. The ablation results demonstrate that these modules collectively mitigate client drift, enhance segmentation accuracy, and strengthen the model's robustness under heterogeneous data distributions.




\begin{figure*}[!htp]
	\centering
\includegraphics[width=\linewidth]{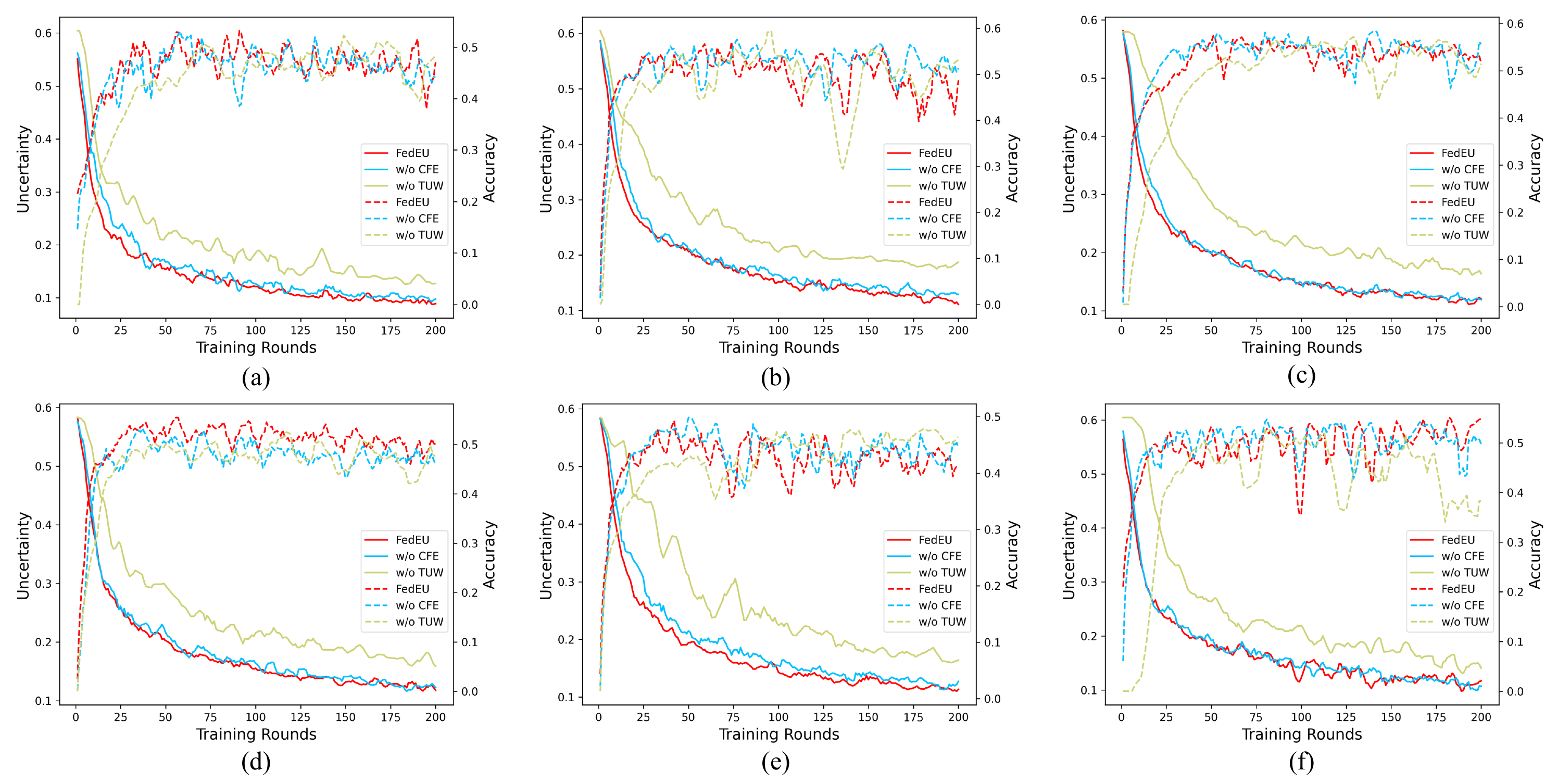}
 \vspace{-8pt}
	\caption{{Training performance with uncertainty constraints using the GF-7 dataset. (a) to (f) are results in the clients of Chongqing, Guangzhou, Lanzhou, Ningbo, Shenzhen and Tianjin, respectively.}}\label{fig_city_uncertainty}
\end{figure*}

\subsection{Parameter Analysis}
To evaluate the effect of evidential uncertainty loss on model performance, we conduct a parameter sensitivity analysis on the weighting factor $\mu$. As shown in Fig.~\ref{para_analysis}, we examine IoU (\%) variations across six clients from different cities, with $\mu$ ranging from 0 to 0.8. The results indicate that FedEU is relatively robust to the choice of $\mu$ within a moderate range (e.g., [0.2, 0.6]), where IoU scores remain stable across most clients. In clients (a), (c), (d), and (f), performance steadily improves or fluctuates slightly within this range, confirming that the uncertainty loss provides consistent benefits without causing instability. Although clients (b) and (e) show more noticeable variation, the impact remains limited and does not significantly degrade segmentation quality. This robustness highlights the practicality of FedEU, as it maintains strong performance without requiring extensive hyperparameter tuning.

\begin{figure*}[!htp]
	\centering
	\includegraphics[width=\linewidth]{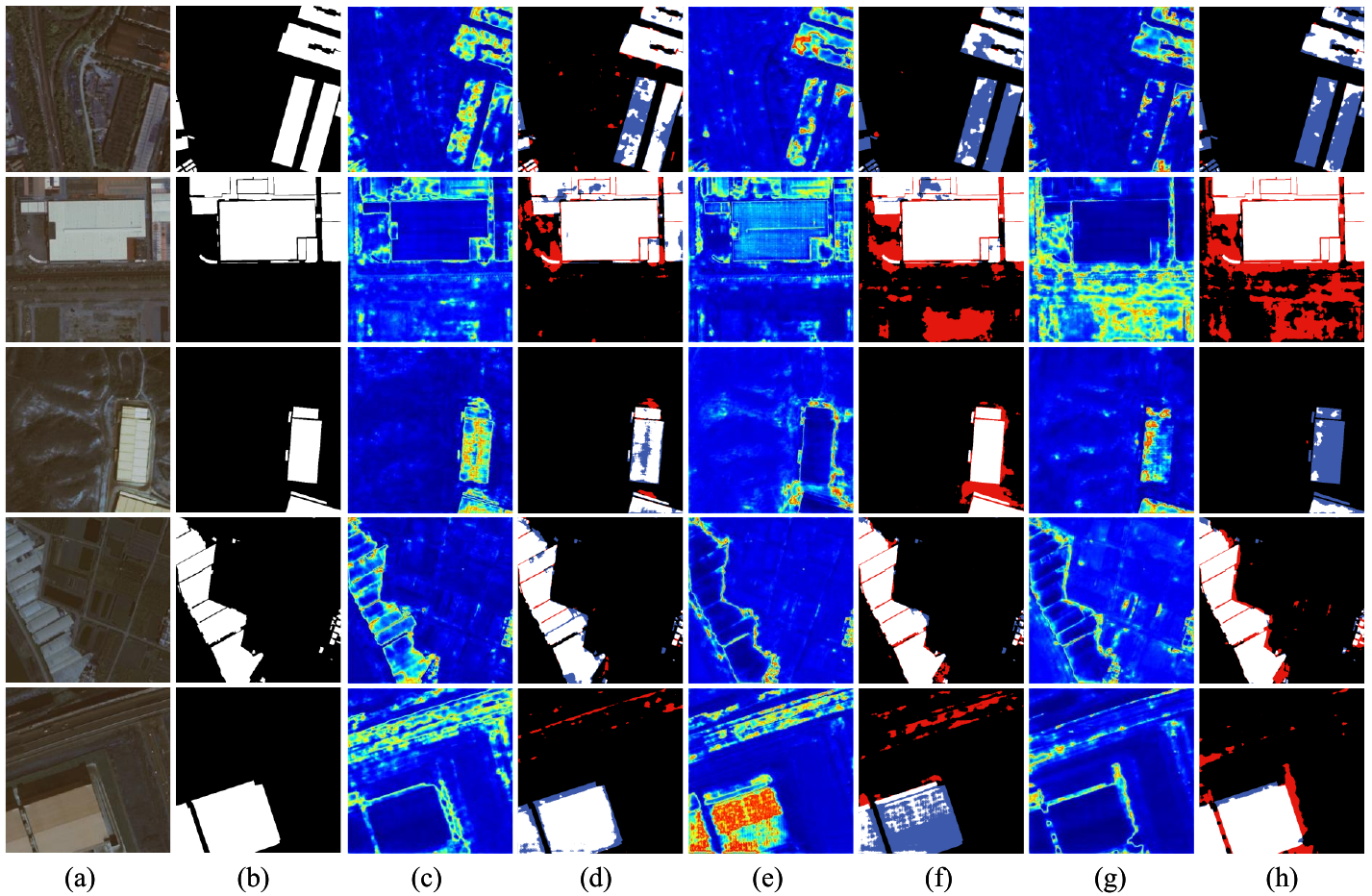}
    \vspace{-8pt}
	\caption{{{Visual analysis of the distributions of evidential uncertainty and prediction errors with different ablations on the GF-7 dataset. (a) Original image. (b) Ground truth. (c)-(d) FedEU. (e)-(f) w/o CFE. (g)-(h) w/o TUW.}}}
	\label{fig_ablation_uncertainty}
\end{figure*}

\subsection{{Uncertainty Analysis}}
\subsubsection{On the Convergence}
To further analyze the impact of the proposed personalization and regularization strategies on model training and uncertainty estimation, we tracked the changes in evidential uncertainty and prediction accuracy throughout the training process. As illustrated in Fig.~\ref{fig_city_uncertainty}, FedEU exhibits a rapid decline in uncertainty at the early training stages, followed by stabilization. It demonstrates a consistent and stable reduction in uncertainty across all clients, and it provides more accurate predictions as the prediction uncertainty decreases.
When the CFE module is removed, the model exhibits slower convergence to the local data distribution. Notably, the absence of the TUW results in consistently higher uncertainty for clients such as Guangzhou, Lanzhou, Shenzhen and Tianjin, caused by unreliable model updates and biased alignment direction under heterogeneous data distributions. Significant fluctuations in prediction accuracy are also observed in clients such as Guangzhou and Tianjin. In contrast, incorporating CFE and TUW constraints promotes faster convergence and more stable prediction performance, and the prediction uncertainties can be reduced more rapidly.
These findings confirm that the proposed FedEU achieves rapid convergence of evidential uncertainty, improves the robustness of uncertainty estimates and ensures stable and balanced model adaptation across diverse clients.


\subsubsection{On the Prediction Errors}
Furthermore, we visualized evidential uncertainty and prediction errors to explore their relations on the GF-7 dataset using FedEU and its variants. As shown in Fig.~\ref{fig_ablation_uncertainty}, higher uncertainties are mainly concentrated inside and along the edges of building entities. 
FedEU and its variations provide reliable uncertainty estimation for model prediction because they leverage maximum entropy theory to effectively capture high uncertainty in unreliable predictions, demonstrating their advantage in uncertainty estimation and reliable decision-making.
Without the CFE module, the model may assign high confidence scores to incorrect predictions in background regions with false alarms, indicating overconfidence in its erroneous outputs. A similar phenomenon is observed in some omitted areas within building entities when the TUM module is eliminated. Meanwhile, the model produces more prediction errors accompanied by higher uncertainties.
Overall, the uncertainty quantification mechanism in FedEU can effectively capture high uncertainty in unreliable predictions and reduce their impacts to improve the model’s adaptability across diverse clients.

\section{Conclusion}\label{sec6}
This paper first explored the remote sensing image segmentation (RSIS) based on the vision foundation model in federated environments by leveraging the parameter-efficient fine-tuning (PEFT) approach to unleash the generalization capability of the vision foundation model, like the Segment Anything Model (SAM), while reducing aggregation and communication overhead.
To achieve this, we proposed FedEU, a novel evidential uncertainty-driven federated optimization method.  The core idea of FedEU is to fully leverage the variations of the model's epistemic uncertainty as constraints to guide the federated optimization.
Specifically, domain-specific evidential uncertainty modeling and a client-specific feature embedding (CFE) module are introduced to enable personalized representation learning and capture local model uncertainties. Furthermore, a Top-$k$ uncertainty-guided weighted (TUW) aggregation strategy is proposed to control the impact of distribution shifts and high-uncertainty predictions on the global model. 
By guiding the local training and global aggregation using evidential uncertainty, FedEU significantly enhances robustness and generalization of the segmentation model under heterogeneous conditions. 


Experimental results demonstrate that FedEU outperforms several state-of-the-art federated optimization methods on three large-scale remote sensing datasets. Ablation studies and uncertainty analyses show that the CFE module effectively captures local data characteristics, while the uncertainty-aware constraints reduce client drift and ensure stable convergence under non-IID conditions. The uncertainty estimations further confirm that FedEU mitigates overconfidence in challenging scenarios and better aligns high-uncertainty regions with prediction errors, thereby making reliable aggregation decisions. This demonstrates the effectiveness of FedEU’s uncertainty estimates and further strengthens the robustness and reliability of the segmentation model.
Future work will focus on extending FedEU to multimodal federated learning and incorporating communication-efficient strategies to enhance scalability and reduce training overhead in large-scale deployments.



%




\ifCLASSOPTIONcaptionsoff
  \newpage
\fi



%
\bibliographystyle{ieeetr}
\bibliography{refer}

\begin{IEEEbiography}[{\includegraphics[height=1.25in,clip,keepaspectratio]{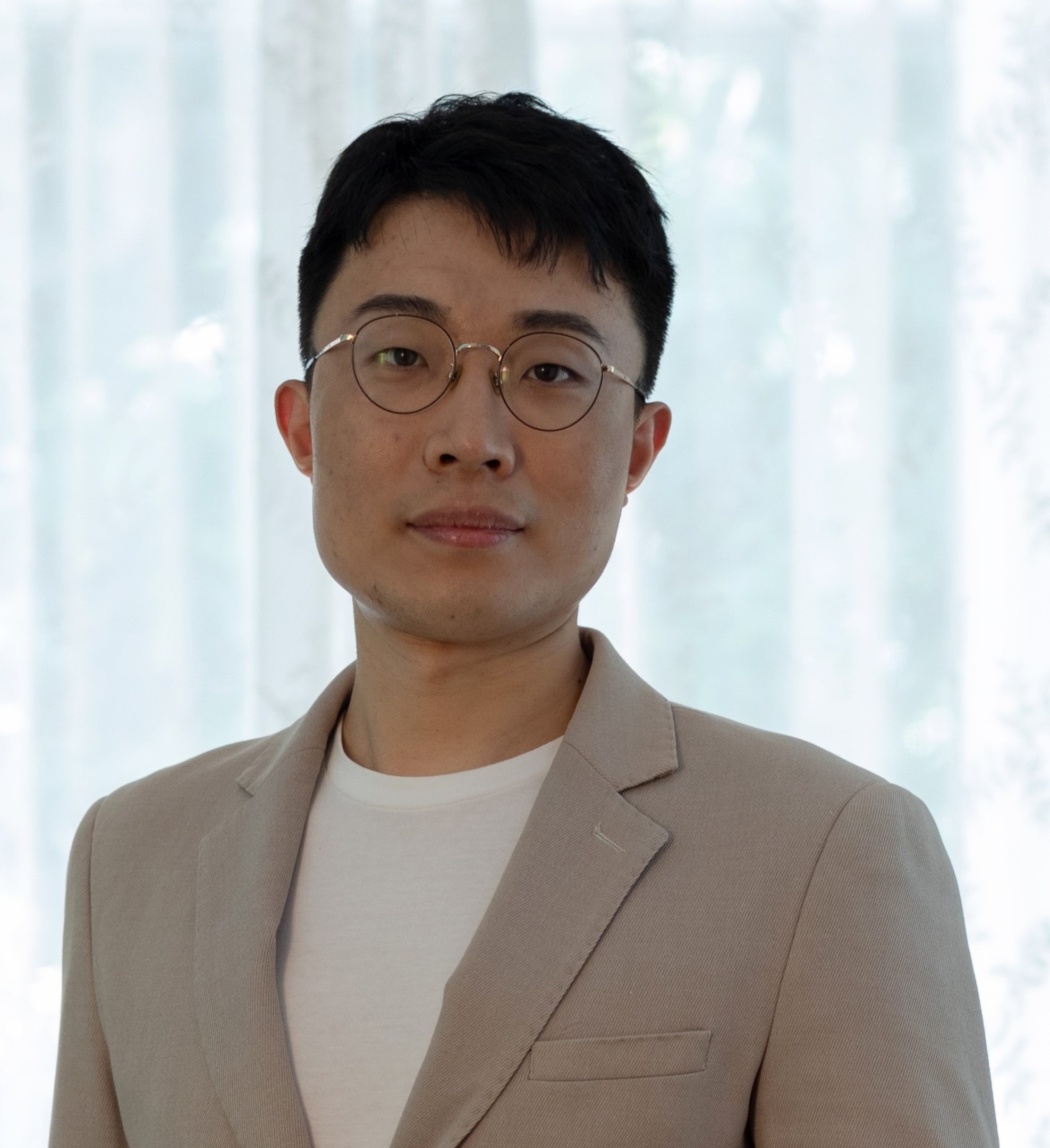}}]
    {Xiaokang Zhang} (Senior Member, IEEE)
    received the Ph.D. degree in photogrammetry and remote sensing from Wuhan University, Wuhan, China, in 2018. 

    From 2019 to 2022, he was a Postdoctoral Research Associate with The Hong Kong Polytechnic University, Hong Kong, and The Chinese University of Hong Kong, Shenzhen, Shenzhen, China. He is currently an Associate Professor with the School of Artificial Intelligence, Wuhan University, Wuhan, China. He has authored or coauthored more than 60 scientific publications in international journals and conferences. His research interests include remote sensing image analysis, computer vision and machine learning. 
\end{IEEEbiography}

\begin{IEEEbiography}[{\includegraphics[height=1.25in,clip,keepaspectratio]{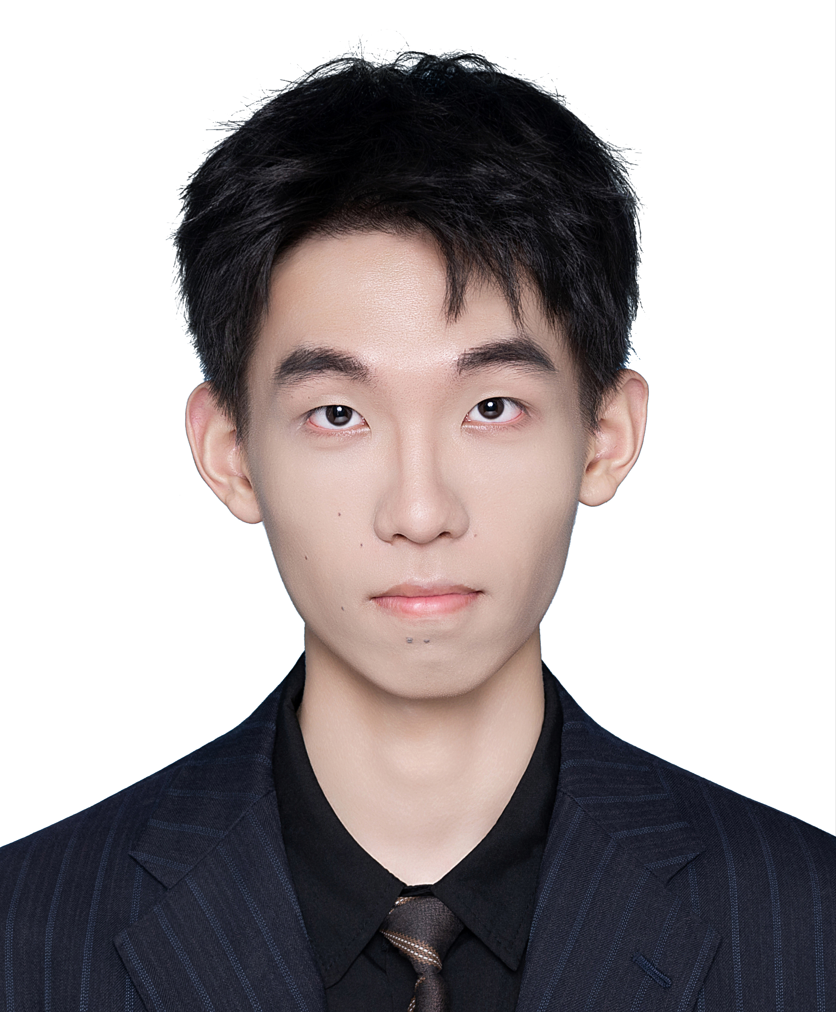}}]
    {Xuran Xiong} 
is pursuing his bachelor’s degree in Electronic Information Engineering at Wuhan University of Science and Technology, Wuhan, China. His research interests include federated learning, semantic segmentation, and remote sensing.

\end{IEEEbiography}

\begin{IEEEbiography}[{\includegraphics[height=1.25in,clip,keepaspectratio]{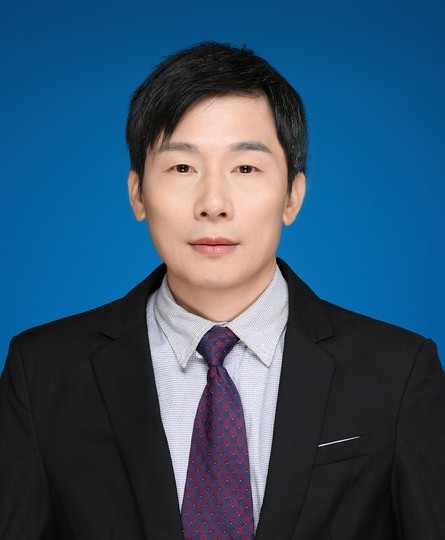}}]
    {Jianzhong Huang} is a senior experimentalist at Wuhan University. His research interests mainly focus on computer systems, intelligent computing systems, and intelligent Internet of Things, with a focus on practical applications and experimental exploration in these fields.
\end{IEEEbiography}

\begin{IEEEbiography}[{\includegraphics[height=1.25in,clip,keepaspectratio]{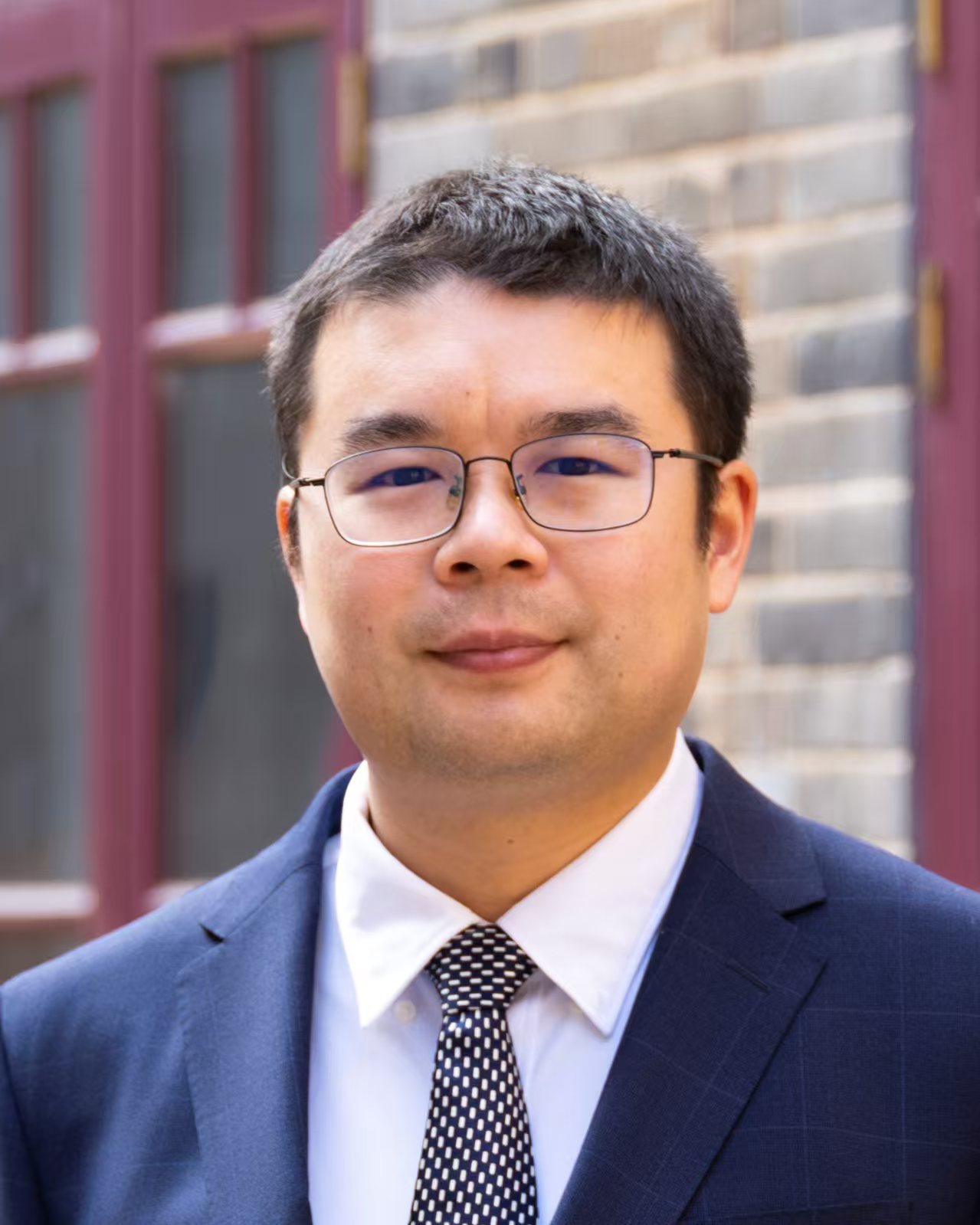}}]
    {Lefei Zhang} (Senior Member, IEEE)
    received the B.S. and Ph.D. degrees from Wuhan University, Wuhan, China, in 2008 and 2013, respectively. He was a Big Data Institute Visitor with the Department of Statistical Science, University College London, U.K., and a Hong Kong Scholar with the Department of Computing, The Hong Kong Polytechnic University, Hong Kong, China. He is a professor with the School of Computer Science, Wuhan University, Wuhan, China, and also with the Hubei Luojia Laboratory, Wuhan, China. His research interests include pattern recognition, image processing, and remote sensing.
Dr. Zhang serves as an associate editor of IEEE Transactions on Geoscience and Remote Sensing and IEEE Geoscience and Remote Sensing Letters.
\end{IEEEbiography}

%









\end{document}